\documentclass[
]{ceurart}

\usepackage{listings}
\begin{document}

\copyrightyear{2023}

\copyrightclause{Copyright for this paper by its authors.
  Use permitted under Creative Commons License Attribution 4.0
  International (CC BY 4.0).}


\title{Chit-Chat or Deep Talk: Prompt Engineering for Process Mining}


\author[1]{Urszula Jessen}[orcid=0000-0002-7282-8451,email=urszula.jessen@gmail.com]
\cormark[1]
\fnmark[1]
\author[2]{Michal Sroka}[orcid=0000-0002-7505-2521, email=msmisiu@gmail.com]
\cormark[1]
\fnmark[1]
\author[3]{Dirk Fahland}[orcid=0000-0002-1993-9363, email=d.fahland@tue.nl]
\cormark[1]
\fnmark[1]

\begin{abstract}
Abstract: This research investigates the application of Large Language Models (LLMs) to augment conversational agents in process mining, aiming to tackle its inherent complexity and diverse skill requirements. While LLM advancements present novel opportunities for conversational process mining, generating efficient outputs is still a hurdle. We propose an innovative approach that amend many issues in existing solutions, informed by prior research on Natural Language Processing (NLP) for conversational agents. Leveraging LLMs, our framework improves both accessibility and agent performance, as demonstrated by experiments on public question and data sets. Our research sets the stage for future explorations into LLMs' role in process mining and concludes with propositions for enhancing LLM memory, implementing real-time user testing, and examining diverse data sets.
\end{abstract}


\maketitle

\section{Introduction}\label{Introduction}

Process mining endeavours, increasingly prominent across various industry domain such as healthcare, manufacturing and supply chains, require stakeholder engagement with diverse skills \cite{reinkemeyer}. This paper propose a novel framework for creating conversational agents capable of directly extracting answers from event data, thus reducing the need for interaction with multiply stakeholders.

Recent advancements in Large Language Models (LLM) have shown great potential in handling tasks such as interpreting nuanced questions from non-experts. Despite the apparent simplicity of their usage, the performance of LLMs is largely reliant on the construction of a well-crafted, nuanced prompt \cite{brown}.
This paper explores the optimal methods for automating tasks typically performed by humans, specifically converting an end-user's question into an event data query. Conventionally, these tasks involve process analysts who comprehend the issues related to process optimization, domain experts who understand the process at hand, and data engineers who can effectively query the background data to generate accurate answers. The key challenge lies in assembling all necessary information to construct the query. By emulating and automating these tasks, this study aims to streamline the workflow and deliver results directly to the end-user.


In this study, we have constructed a framework, supported by a Large Language Model (LLM), which emulates the tasks and skills associated with various process mining participants. We have developed prompts that emulate individual roles and used an orchestrator to integrate them. Finally, we employed the corpus of process mining questions collated by Barbieri et al. \cite{barb1} and evaluated our framework using the BPI Challenge 2019 dataset \cite{dongen}.Our findings indicate that in 77\% of instances, LLMs were capable of fully or partially comprehending the question and outlining the appropriate solution. Furthermore, in 68\% of cases, the model provided either the correct or a partially correct answer.

This paper is organized as follows: Following this introduction, Chapter 2 provides an overview of the background and relevant literature in the field. Chapter 3 describes our proposed approach for integrating Large Language Models (LLMs) into conversational interfaces for process mining. Chapter 4 then illustrates an experimental evaluation to confirm the effectiveness of using LLMs for conversational querying in process mining.

\section{Background and relevant literature}\label{Background}

Process mining a growing branch of data science, enables companies to analyze their business processes based on their event data. It provides insights into prevalent performance bottlenecks, inefficiencies, and compliance risks in operational work. Nonetheless,the  discovered data models provide often complicated, underfit or Spaghetti-like diagrams\cite{aalst}, and the underlying event-log data can be challenging to analyze through non-experts\cite{barb1}. 

This study aims to address this challenge by proposing a novel approach: integrating large language models (LLMs) into conversational interfaces for process mining querying. A large language model (LLM) is a natural language processing tool(NLP) that can understand and generate human-readable text\cite{johnson}. This approach provides users, without technical or process-mining expertise, direct access to the insights derived from their event data.

The success of many process mining projects often relies on effective interaction between diverse stakeholders\cite{brocke}. The complexity of data and results, coupled with communication requirements, has amplified the interest in enhancing usability and understandability for all participants \cite{martin, kubrak}
Extending this, Dumas et al. suggest the need for not only improved usability but also the development of an Augmented Business Process Management System. This system would facilitate a conversationally actionable interface between humans and IT systems \cite{dumas:fourn}.



Recently, large language models (LLMs) such as GPT-4 have shown remarkable progress in natural language processing tasks. They have demonstrated their capability in generating human-like responses to complex queries and providing accurate answers to questions \cite{choi}. LLMs learn substantial linguistic and factual world knowledge from a vast corpora of data. They can execute multiply task, but their performance can be highly variable and in some cases unsatisfactory\cite{taylor}. The quality of the factual information obtained from the LLM depends on carefully designed and nuanced prompts\cite{takeshi}. To address this, prompt engineering has been proving success \cite{brown}.Prompt engineering in the context of large language models involves designing and optimizing prompts to evoke desired responses from these models. In essence, prompts serve as input texts that direct the language model to produce specific outputs. The efficacy of prompt engineering has been demonstrated across a range of applications. These include knowledge-based question answering \cite{yang}, essay writing \cite{sezgin}, sentiment analysis \cite{sezgin}, and addressing medical challenge problems \cite{harsha}. 
This study investigates the feasibility of using Large Language Models (LLMs) for conversational querying in process mining. It examines the challenges inherent to natural language processing (NLP) and explores strategies to mitigate some of these challenges, notably through prompt engineering and task orchestration based on varying outcomes.
\section{Architecture and process for prompt engineering}\label{approach}
\subsection{Generic approach}
One of the main challenges in implementing natural language interfaces for process mining analysis using LLMs is the creation of nuanced, context-specific task descriptions. These must generate SQL queries that are not only semantically correct but also precisely address the user's question.
A common strategy employed to address such problems entails the generation of a prompt, including the users question, followed by a request to generate SQL which answers the question. Such prompt typically includes instructions regarding the format and characteristics expected in the SQL statement. The SQL is then executed against an event log database and the resulting solution, or an error, is fed back to the system which decides whether to display it to the user or withhold it based on predefined criteria.
This approach, however, underperforms in the following scenarios: 
\begin{enumerate}
\item The question contains domain-specific terms.
    \begin{enumerate}
        \item Effect: LLM is unable to understand the specific terms and falls back to generic meaning of the words. Often this leads to incorrect interpretation of the questions and therefore answer is inappropriate.
    \end{enumerate}
\item The SQL is malformed.
    \begin{enumerate}
        \item Effect: Error in execution of SQL, the user does not see an answer.
    \end{enumerate}
\item The question is complex and requires multiple SQL statements.
    \begin{enumerate}
        \item Effect: The answer is malformed or meaningless SQL, alternatively, the generated SQL makes many assumptions and does not answer the actual question. 
    \end{enumerate}
\item The data model is not standard or LLM does not have sufficient information about its structure.
    \begin{enumerate}
        \item Effect: the LLM hallucinates column names, makes assumptions about data formatting. The ensuing SQL statements either exhibit functional inadequacies, leading to erroneous outcomes, or generate outputs characterized by nonsensical or stochastic properties.
    \end{enumerate}
\end{enumerate}

In the subsequent sections, we present an architectural framework designed to effectively address the aforementioned limitations.

\subsection{Architecture}
To illustrate the architecture of our solution, consider the following scenario: we want an LLM to translate the question "What is the main bottleneck in my process in department A?" into a SQL query. The task requires specific information to execute. Necessary elements include:

\begin{enumerate}
\item The data structure, such as the case\_concept\_name field, timestamp, or event field names.
\item Process-mining-specific information, such as the information on how the eventlog is built what is the meaning of case, activity or timestamp in that context
\item Domain-specific information, such as the interpretation of the term "process bottleneck", and instructions on how it is related to the underlying data. Also as bottleneck can mean different things such as resource or time bottleneck.
\item Data set-specific information, such as the method for calculating duration if there are no \_start, \_end timestamps for each activity.
\item Mapping of data to domain knowledge e.g. what is the representation of department A in the given data set.
\end{enumerate}
\vspace{0.2cm}
\begin{figure}
\includegraphics[scale=0.6]{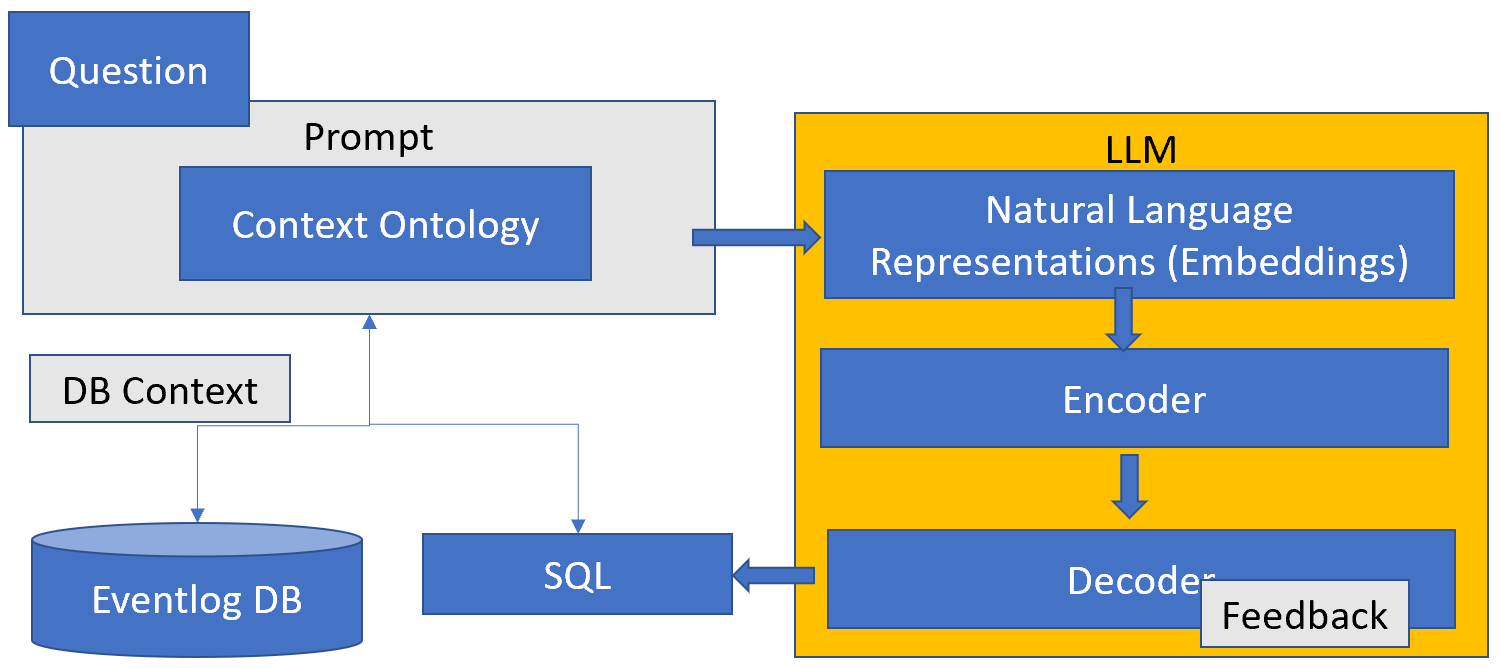}
\vspace{0.2cm}
\caption{Architecture of a conversational agent for process mining }
\end{figure}


Figure 1 depicts the overall architecture of our framework. To generate a response, a customized \textit{prompt}, integrating both general guidance on output structure and specific context, is created for each query. This necessitates an update to the eventlog \textit{DB context} (field names, structure, data types) for each new query. The LLM then establish a list of required information and employs the \textit{context ontology} to enhance the prompt. Upon receiving the prompt, the LLM crafts an \textit{SQL query} to be run against \textit{the Eventlog in the DB}. If the execution is successful, the user is provided with a response; if it fails, the LLM is given \textit{feedback}—previous results, the SQL error, and instructions for correction. This iterative cycle continues until a satisfactory answer is achieved or a set loop limit is met.

\subsection{Conceptualizing Interaction with LLM as a Process}

In the context of process mining, conversational agents face a significant challenge: they must effectively integrate diverse skill sets and knowledge domains to optimize company-specific processes. This requires data engineering expertise for understanding data contexts, process analysis capabilities to comprehend terms like "process model," "variant," "deviations," and to identify bottlenecks and improvement opportunities. Moreover, responding to domain-specific queries may require domain expert's knowledge.
\begin{figure}
\includegraphics[width=\linewidth]{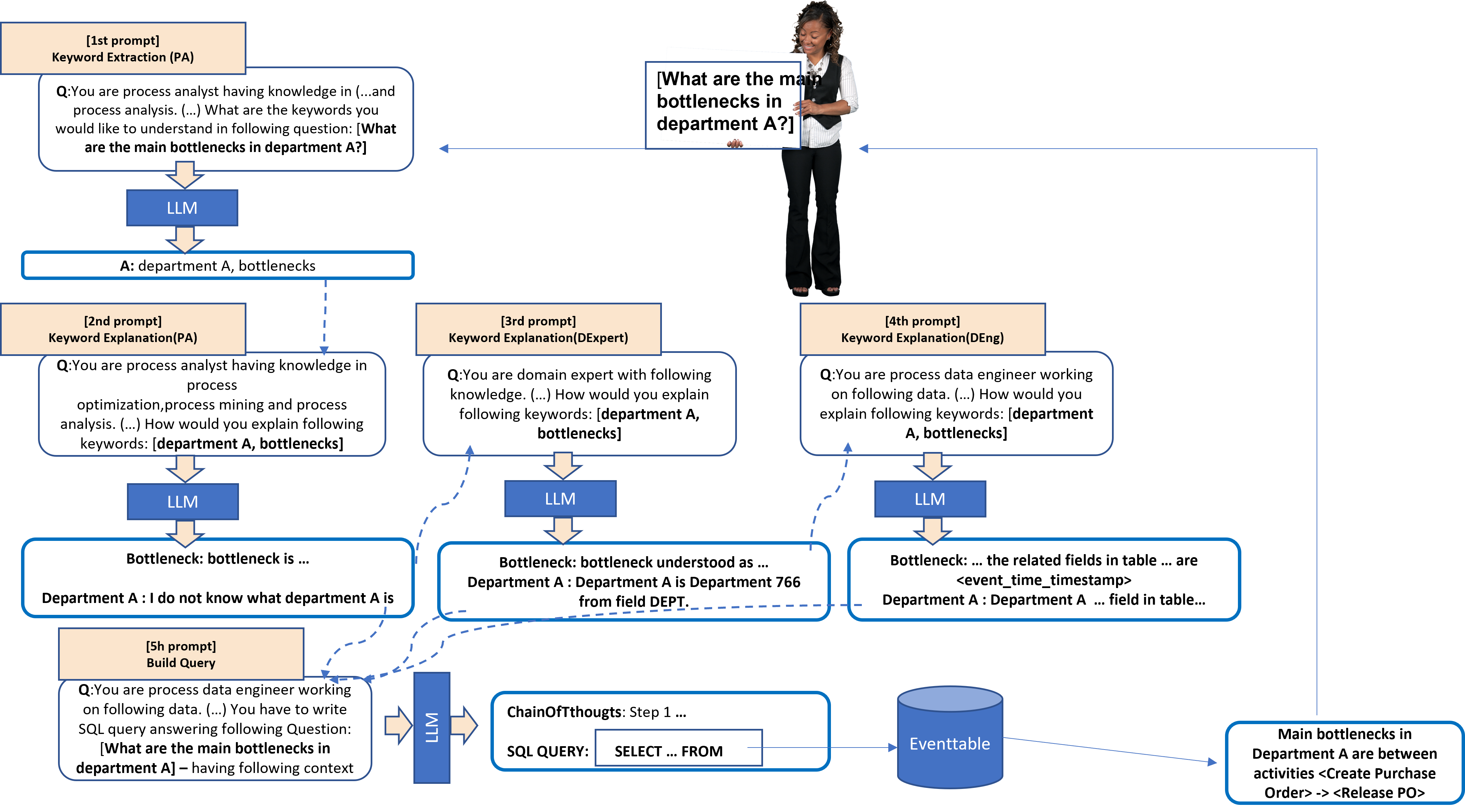}
\vspace{-0.3cm}
\caption{The general process of prompt engineering. }
\end{figure}

Figure 2 illustrates the process we have developed to address the diversity of required skills and knowledge. One of the unique attributes of LLMs is their effectiveness in understanding problems when specific keywords are activated, such as "you are an experienced data engineer with expertise in databases and SQL queries (...)." By establishing a process and assigning roles to segregate duties and context, enabling the LLM to focus on its specific task, the outcomes of multiple queries present a greater specificity. In order to answer a specific question a range of different prompts has to be created in order to query different aspects of a question. 
Once the context has been established, the initial prompt to the Data Engineer can be formulated. This prompt would only contain the necessary data and would instruct the LLM to construct the SQL query and elucidate the reasoning behind the sequential steps of obtaining an answer to the user's query\footnote{ Additional prompt examples can be found at \url{https://tinyurl.com/chitchatdeeptalk}}.

The diagram in Figure 3 outlines the general process constructed in our proposed 
framework.
Upon receiving a user's question, the application (orchestrator) first checks for similar questions in the database. These questions have previously been submitted to an LLM to generate embeddings for each of them. If any of these vectors show a similarity greater than 0.9 with the new question, the orchestrator verifies the success of the previous answer. If successful, the sql query is executed and answer is immediately forwarded to the user. Otherwise, the orchestrator forwards the execution to prompt creation task where the context of different perspectives of the question is assembled into one general, tailored to the question prompt.
\begin{figure}
\includegraphics[scale=0.6]{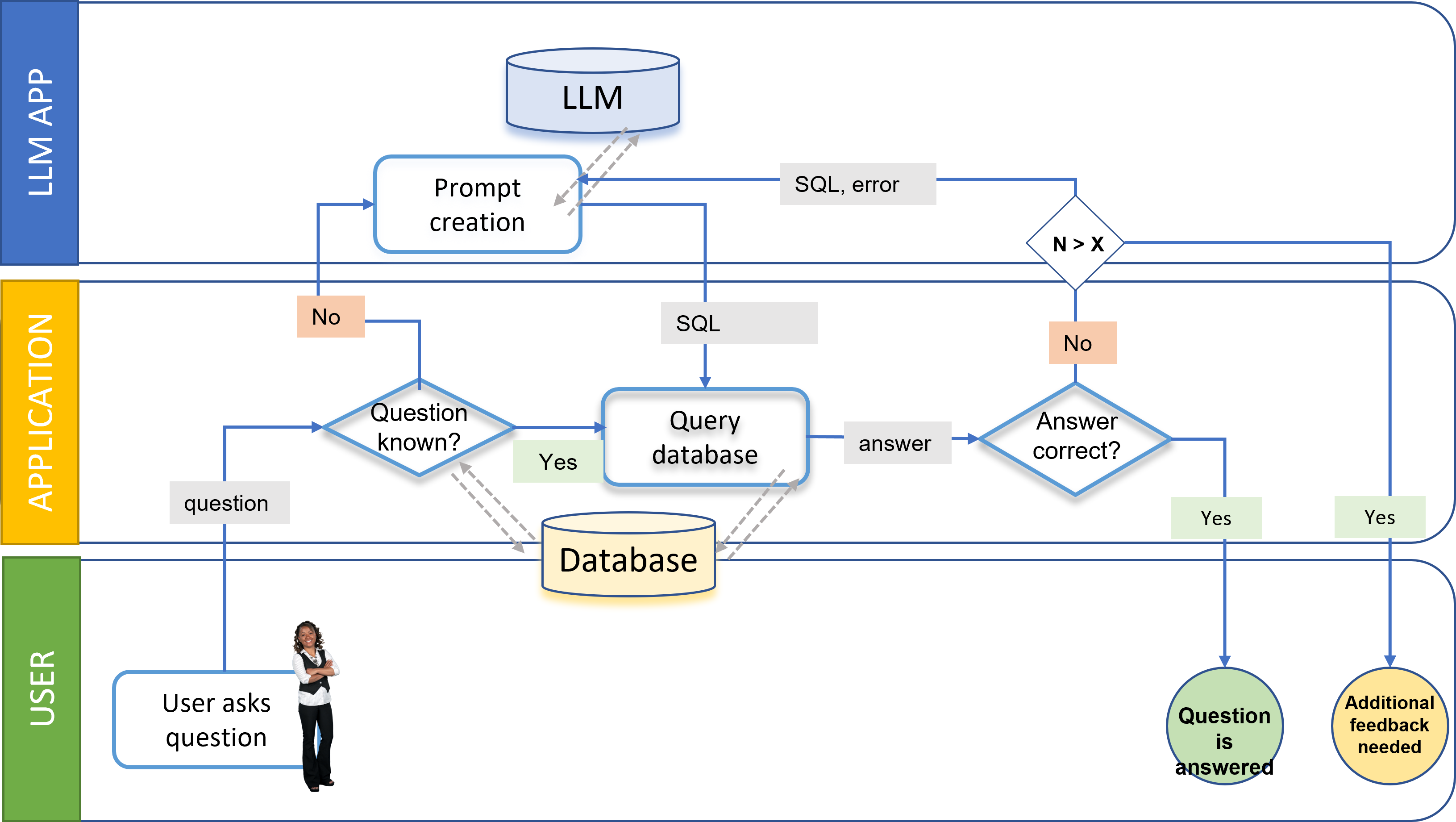}
\vspace{-0.3cm}
\caption{Interaction between user and LLM as a process}
\end{figure}

Upon crafting the SQL query, it would be dispatched to the database via the orchestrator. In the event of errors, the Data Engineer prompt would be supplemented with the database's error information and executed again. Upon successful query execution, the user would receive an answer to their question. After two loops the model would be changed from GPT 3.5-turbo to GPT-4. The prompt creation component can also receive context from the most similar questions from the database. If it is not possible to create correct answer, the orchestrator would ask the user to provide additional feedback to assist in answering the question.

\section{Evaluation}\label{evaluation}
We developed our proposed conversational agent architecture and prompt engineering process using Python, employing GPT-3.5-turbo and GPT4 as Large Language Models (LLMs). We conducted a study to assess the effectiveness of this approach, employing a real-world dataset and real-life questions.
\subsection{Data used in study}

Questions in Barbieri et al.'s work\cite{barb1, barb2} were sorted into four categories: process model, event log data, analysis, and advanced analysis, across 23 perspectives like bottleneck analysis and conformance checking. The question corpus, primarily in Portuguese and translated by volunteers, included malformed or unrelated queries. Rather than excluding these flawed questions,as was done by Barbieri et al.\cite{barb1, barb2}, we tested them all on the LLM, adjusting only year-specific queries to "2019". This was done to explore the LLM's potential in comprehending complex or unclear texts, where traditional rule-based systems may struggle.The questions has been asked against the BPI Challenge 2019 dataset \cite{dongen}\footnote{The dataset is a log of a the purchase order handling process for a large multinational company operating from the Netherlands in the area of coatings and paints.  The log contains a total of 1,595,923 events spread over 51,734 cases. The cases are created at the level of the position item of a purchase order.There are 42 different events in the eventlog. The events were executed mostly in 2018, 2019 and cases are spread over 13,881 variants.}.

\subsection{Initial Experiment Results}
We executed the experiment in two stages. The initial round, focusing on the corpus's first 100 questions, was designed to identify and address weaknesses in our methodology. After rectifying these, we progressed to the second round, evaluating the entire corpus.
During the first round, we also fine-tuned our prompts and refined the architecture to minimize human intervention. Insights gained from this preliminary round were instrumental in enhancing our architectural approach for the second round involving the full corpus of questions. Details of these modifications and the enhanced process are discussed in Section 3.

\subsection{Adjusted setup and final experiment results}
To handle the complexity of defining expected answers, we manually evaluated the responses using certain criteria, dividing correct answers into \textbf{fully answered} and \textbf{partially answered} categories.

\begin{itemize}
\item A \textbf{fully answered} question is one that correctly addresses the question, including any additional information provided by the LLM\footnote{For example, in response to a question like "Which tasks have the maximum duration," we would consider the LLM's answer correct if it not only identifies the tasks with the maximum duration but also provides additional information such as the specific duration of these tasks.}.
\item \textbf{Partially answered} questions are those where the executed database query didn't generate errors and the resulting answer required some expert interpretation\footnote{For instance, a query requesting the top variants in a process model might return all variants sorted by frequency, but fail to specify the top 3, 5, or 10 variants. These types of answers, which could be refined with extra feedback or contextual information for the LLM, were considered partially answered.}.
\end{itemize}

In contrast, \textbf{wrong answered questions} encompassed those that did not generate any code (resulting in SQL Server errors), as well as responses that, while calculated, could not be considered correct. 

In our evaluation process for the responses generated by the Large Language Model (LLM), we identified two primary criteria: \textbf{Understood} and \textbf{Partially Understood}. These criteria were assessed based on the logical chain of thought demonstrated by the model in its responses.
\begin{itemize}
\item
An \textbf{Understood} question is distinguished by the LLM's comprehensive and accurate response, suggesting a full grasp of the query. This entails proper problem identification and subsequent application of relevant operations to derive a solution\footnote{For instance, when asked to identify the primary bottlenecks in a process, the LLM correctly pinpoints them by measuring the durations between events, identifying cases that exceed the average duration, and ultimately, providing the specific bottleneck events. This line of reasoning displays a thorough understanding of the question.}.
\item Conversely, a \textbf{Partially Understood} question is characterized by the LLM's partial grasp of the question or its incomplete or partially correct response\footnote{A prime example is the query, "How many services do not conform to the online sales model?" Here, while the LLM rightly perceives the necessity to contrast event sequences with a predefined online sales model, it may fall short in comprehensively determining what qualifies as conformance to the model, or may not produce a complete count of non-conforming services. This yields a response that only partially addresses the question.}.
\end{itemize}
Our findings indicate that the Chain-Of-Thought, particularly for Understood questions, is not just valuable for end-users but also enlightening for process analysts. It sheds light on how to break down complex questions into manageable parts and solve them systematically.

For the examples of "Understood" and "Partially Understood" questions with LLM reasoning behind, refer to the file in the provided folder at \url{https://tinyurl.com/chitchatdeeptalk}.

Table \ref{table:results} summarizes the results as how many of the 795 questions were (partially) answered or understood by the agent compared to the results of Barbiere at al.\cite{barb2}.

\begin{table}[h!]
\centering
\begin{tabular}{ccccc}
\toprule
\textbf{Result}      & \textbf{Count} & \textbf{Ratio} & \textbf{Count\cite{barb2}} & \textbf{Ratio\cite{barb2}} \\
\midrule
Answered             & 285            & 36\%           & 266               & 56\%              \\
Partially answered   & 254            & 32\%           & 42                & 9\%               \\
\midrule
\midrule
Understood           & 155            & 19\%           & 304               & 64\%              \\
Partially understood & 459            & 58\%           & 42                & 9\%               \\
\midrule
\bottomrule
\end{tabular}
\caption{Our experiments were conducted on a corpus of 795 questions, a substantial increase from the 476 questions evaluated in the study by Barbieri et al.\cite{barb2}. These findings present a comparison between our results and those from the previous research.} 
\label{table:results}
\end{table}

Though the proportion of fully or partially answered questions didn't increase significantly, utilizing the LLM permitted the assessment of a larger corpus of questions. 
In contrast to the method employed by Barbieri et al.\cite{barb2}, which solely examined whether the model comprehended the semantic essence of the question, our study additionally assessed the model's capacity to formulate logical reasoning for the correct solution. This approach provides a more in-depth perspective on the understanding capabilities of the LLM.
For future research, an important objective would be to explore ways to effectively measure the understanding of LLMs to enhance the results.

\subsection{Improving the results with process orchestrator and one/few shot learning}
In the large language model the model is already trained on the vast amount of data and it is not possible to adjust the model in order to increase the performance of the model in specific tasks. In the case of such models a set of methods has been developed in order to fine-tune the model to the function it should fulfill. One of the methods is already explained in the Approach Chapter and contains of asking the LLM to explain the reasoning behind the answer Chain-Of-Thought. Similarly the additional methods used to fine-tune such models is zero- and few-shot learning. \footnote{In the case of LLMs, zero-shot learning is achieved by providing the model with a prompt that describes only the task it needs to perform. The model then generates an output based on its understanding of the prompt and its pre-existing knowledge. Few-shot learnings achieved by providing the model with one or a few examples of the task it needs to perform, along with a prompt that describes the task. The model then generates an output based on its understanding of the prompt and the provided examples \cite{takeshi}.}.

\begin{table}[h!]
  \centering
    \begin{tabular}{lrr}
      \toprule
     Zero vs Few Shot & GPT 3.5 & GPT 4.0 \\
      \midrule
      Zero Shot & 49 & 0 \\
      Few Shot & 12 & 193 \\
      \textbf{Sum (Partially Answered)} & \textbf{61} & \textbf{193} \\
      \addlinespace 
      Zero Shot & 61 & 0 \\
      Few Shot & 46 & 178 \\
      \textbf{Sum (Fully Answered)} & \textbf{107} & \textbf{178} \\
      \addlinespace 
      \textbf{Sum (Partially and Fully Answered)} & \textbf{168} & \textbf{371} \\
      \bottomrule
    \end{tabular}
  \caption{Comparison of GPT 3.5 and GPT 4.0 performance}
  \label{table:gpt_comparison}
\end{table}
In our procedure, we utilized various methods to enhance the LLM's responses. The results of our experiments are displayed in Table \ref{table:gpt_comparison}. Our application's orchestrator attempted to execute the SQL code after each LLM response. In 61 instances, the GPT 3.5-Turbo Model produced the correct answer without additional shots (Zero Shot) and in 49 the answer was partially correct. If the code failed to execute, we asked the model to correct the error, supplying it with the error message. This method further improved 46 (Partially Answered) and 12 (Fully Answered) responses with GPT 3.5-Turbo.If these attempts still failed, the orchestrator involved the GPT-4 model, sending the entire conversation to it. If this approach did not yield the desired outcome, the orchestrator provided example code or generated additional context (one or few-shot). This improved an additional 61 cases in total for GPT 3.5-Turbo.  This managed to enhance 193 partially correct responses for GPT 4.0 in few-shot mode, and a further 178 fully answered questions.

It's evident from these results that GPT-4 performs best with few-shot learning.  However, the significant cost associated with this model should not be overlooked. The GPT-3.5 Turbo model was capable of producing satisfactory results in 168 cases. The overall cost factor must be a part of the evaluation as well; for our case of answering approximately 800 questions, the total expenditure was around \$60. In a real-world scenario with live data, this could translate to considerably higher costs. Therefore, the balance between accuracy and cost is essential when choosing between these solutions.

\section{Conclusion}\label{conclusion}

In this paper, we've delved into the potential of Large Language Models (LLMs) in enhancing conversational agents for process mining. We've proposed a framework that fosters data-focused conversations and highlighted techniques that may boost model performance.

We've demonstrated the value of incorporating supplementary data to improve LLM outcomes but underscore that this field still holds considerable untapped potential. We suggest future research could explore the idea of external memory for LLMs to retain context over extended interactions, and investigate the effectiveness of new prompt engineering methods.

To further progress conversational agents, understanding the concept of "understanding" is critical. Detecting answers that are technically correct but semantically incorrect is another challenge, referred to as 'hallucination', that needs to be addressed.

Live system testing with real users could provide valuable insights into real-world effectiveness. Moreover, studying system responses to varied user interactions could highlight its resilience and versatility. Examining additional datasets within the process mining domain may also yield unique challenges and insights, potentially refining our framework further.

In conclusion, our research paves the way for harnessing LLMs in process mining, aiming to alleviate some barriers for non-expert users. We aspire that our work will inspire continued exploration in this area, leading to more intuitive, accessible, and efficient process mining tools.

\begin{acknowledgments}
  Thanks to the developers of ACM consolidated LaTeX styles
  \url{https://github.com/borisveytsman/acmart} and to the developers
  of Elsevier updated \LaTeX{} templates
  \url{https://www.ctan.org/tex-archive/macros/latex/contrib/els-cas-templates}.  
\end{acknowledgments}

\bibliography{bib}

\appendix

\section{Online Resources}

The additional sources with prompt examples are available under
\begin{itemize}
\item \href{https://tinyurl.com/chitchatdeeptalk}{https://tinyurl.com/chitchatdeeptalk}

\end{itemize}

\end{document}